\documentclass{article}

\usepackage{PRIMEarxiv}
\usepackage{makecell}
\usepackage[utf8]{inputenc} 
\usepackage[T1]{fontenc}    
\usepackage{hyperref}       
\usepackage{url}            
\usepackage{booktabs}       
\usepackage{amsfonts}       
\usepackage{nicefrac}       
\usepackage{microtype}      
\usepackage{lipsum}
\usepackage{fancyhdr}       
\usepackage{graphicx}       
\graphicspath{{media/}}     
\usepackage{amsmath}
\usepackage{subfig}
\usepackage{float}
\usepackage{multirow}
\usepackage{algorithm2e}
\title{Generation of artificial facial drug abuse images using Deep De-identified anonymous Dataset augmentation through Genetics Algorithm (3DG-GA).
\thanks{\textit{\underline{Citation}}: 
\textbf{Authors. Title. Pages.... DOI:000000/11111.}} 
}

\author{
  Hazem Zein \\
  LISSI Laboratory \\
  Université Paris-Est Créteil \\
  F-94400 Vitry-sur-Seine, France\\
  \texttt{hazem.zein@u-pec.fr} \\
   \And
  Lou Laurent \\
  Faculty of Science and Technology \\
  Université Paris-Est Créteil \\
  Créteil, 94000, France\\
  \texttt{lou.laurent@etu.u-pec.fr} \\
   \And
  Régis Fournier \\
  LISSI Laboratory\\
  Université Paris-Est Créteil \\
  F-94400 Vitry-sur-Seine, France\\
  \texttt{rfournier@u-pec.fr} \\
     \And
  Amine Nait-Ali \\
  LISSI Laboratory \\
  Université Paris-Est Créteil \\
  F-94400 Vitry-sur-Seine, France\\
  \texttt{naitali@u-pec.fr} \\
}

\begin{document}
\maketitle

\begin{abstract}
In biomedical research and artificial intelligence, access to large, well-balanced, and representative datasets is crucial for developing trustworthy applications that can be used in real-world scenarios. However, obtaining such datasets can be challenging, as they are often restricted to hospitals and specialized facilities. To address this issue, the study proposes to generate highly realistic synthetic faces exhibiting drug abuse traits through augmentation. The proposed method, called "3DG-GA", Deep De-identified anonymous Dataset Generation, uses Genetics Algorithm as a strategy for synthetic faces generation. The algorithm includes GAN artificial face generation, forgery detection, and face recognition. Initially, a dataset of 120 images of actual facial drug abuse is used. By preserving, the drug traits, the 3DG-GA provides a dataset containing 3000 synthetic facial drug abuse images. The dataset will be open to the scientific community, which can reproduce our results and benefit from the generated datasets while avoiding legal or ethical restrictions. 
\end{abstract}


\section{Introduction}\label{sec1}
The ability to create realistic synthetic images has successfully been achieved in recent years. In particular, Computer-generated images, commonly called synthetic images, can produce images of high realism. For instance, synthetic digital faces can be created artificially but appear to be photographs of real people. Within this context, Adversarial Networks (GANs) were used to reach this level of realism by producing synthetic images when training models over large datasets of authentic images. Among others, StyleGAN2\cite{karras_analyzing_2020} is a popular model used to create high-resolution (1024x1024) images of human faces, including people of different genders, ages, etc.

Some studies have considered synthetic image generation of MRI \cite{arita_synthetic_2022} \cite{frid-adar_gan-based_2018}, retinal \cite{chen_deepfakes_2021} \cite{coyner_synthetic_2022}, and facial skin disease \cite{zein_generative_2021} images related to healthcare and medical applications. The corresponding databases are generated using GANs, requiring relatively large training images. This constraint makes GANs ineffective if data access is limited (e.g. privacy concerns). In this study, we propose to produce highly realistic synthetic faces exhibiting drug abuse traits from a small database. Specifically, we consider the Illicit Drug Addicts database (IDAD) \cite{faceofmeth}, containing images of individuals addicted to methamphetamine. It shows the impact of drug use on facial appearances \cite{nida._2019} \cite{freese_effects_2002} \cite{noauthor_methamphetamine:_2012}, such as:

\begin{itemize}
    \item Eyes: dark circles under the eyes,
    \item Nose: nose irritation due to bleeding and runny nose,
    \item Mouth: tooth decay due to the corrosive nature of the drug, which leads to increased infections and swelling of the face,
    \item Skin: Sores and severe acne due to itching and injury,
    \item Wrinkles: Drug-induced loss of appetite leads to weight loss.
\end{itemize}

Therefore, this work focuses on this issue by proposing a method (3DG-GA) to generate a synthetic database of realistic and anonymous faces of drug users. 3DG-GA, is Genetic Algorithms (GA) based, including GAN artificial face generation, face fusion, face forgery detection, and face recognition. The anonymization of the data prevents any potential concerns related to privacy aspect. This paper is organized as follows: in section 2, the presentation of the database, a definition of the genetic algorithm and the explanation of the method used to create the database will be discussed. Section 3 will discuss the results obtained in this work. Finally, section 4 will conclude this work.

\section{Materials and Methods}\label{sec2}

\begin{figure}[H]
    \centering
    \includegraphics[width = \textwidth]{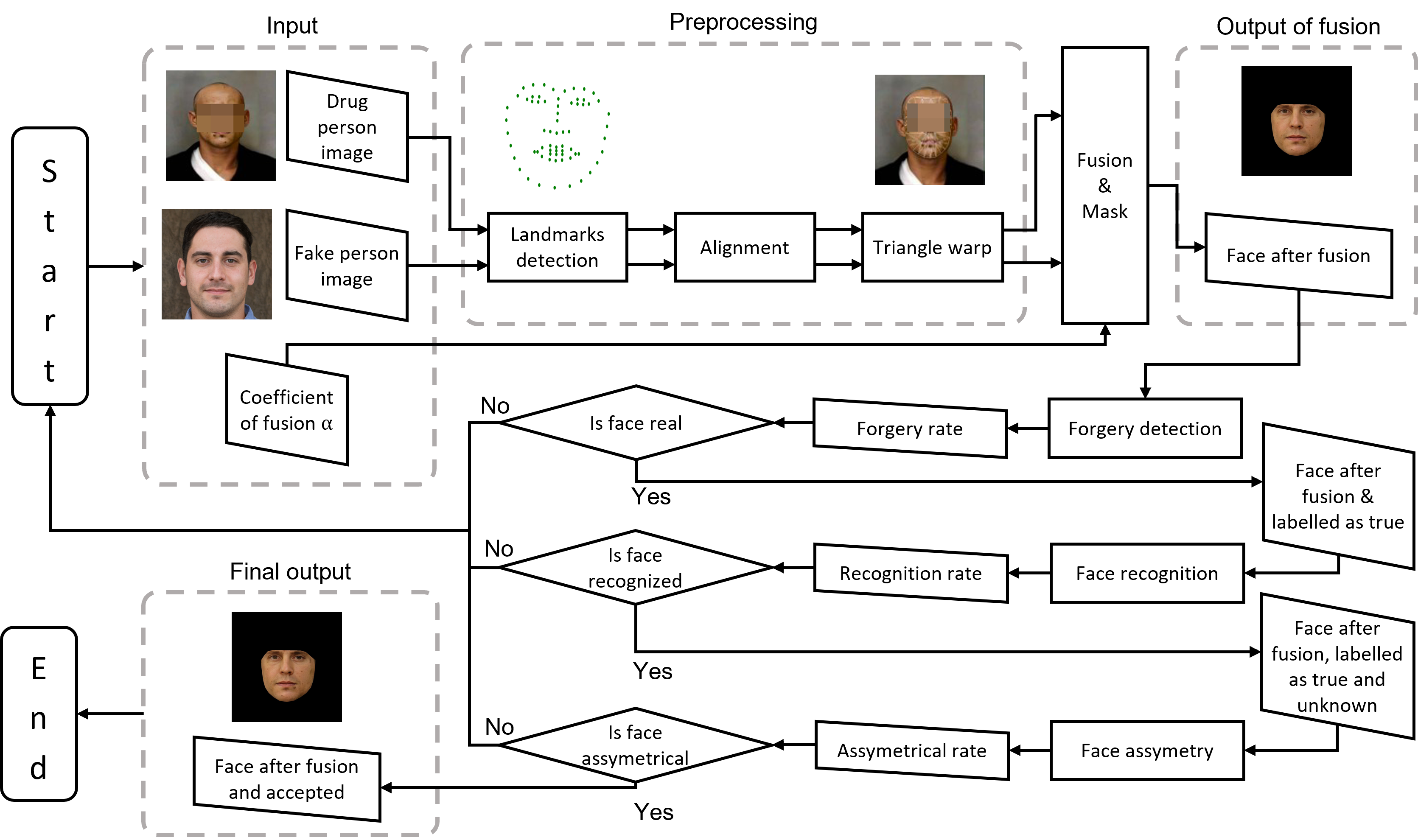}
    \caption{Flowchart of the database generation methodology used during this study.}
    \label{fig:methodology}
\end{figure}

The methodology used during this work is composed of 7 steps shown in Fig. \ref{fig:methodology}. The goal is to realize a fusion of two faces in order to obtain a mixture of these two faces creating a face of a drug addict, realistic and anonymous. This merging takes as input images:

\begin{enumerate}
    \item A healthy subject generated with StyleGAN2 \cite{karras_analyzing_2020} or a drugged person generated by the described method.
    \item A drug addict from the initial database \cite{faceofmeth} or a drug addict generated by the method.
\end{enumerate}

The steps are as follows:

\begin{enumerate}
    \item Landmarks detection: The 68-landmarks-extraction from Dlib's implementation \cite{kazemi_one_2014} is used.
    \item Alignment: The landmarks detected on the two input images are aligned to match the two sets.
    \item Triangle warping: Creation of an identical triangular mesh on the two aligned images.
    \item Merge: Based on the mesh created in the previous steps, the merge is performed by interpolating the two faces. The fusion can be parameterized using a coefficient called $\alpha$: This coefficient represents the percentage used of the drugged face during the fusion. For example, for an $\alpha = 0.5$, the fusion will be $50\%$ for the healthy face and $50\%$ for the drug face. An $\alpha = 1$, means that the fusion is done only with the drugged face and an $\alpha = 0$ is a fusion with the healthy face only. 
    \item Forgery detection: This allows, after the generation of a face by fusion, to check if this image is real or not by providing a percentage. The closer the result is to $100\%$, the more the image will be labeled as real. The details of the method are described in section 2.4
    \item Face recognition: In order to guarantee the anonymity of the generated data, we use a face recognition algorithm explained in section 2.3. The recognition is performed between the face features of the generated image and the face features of the initial database.
    \item Face asymmetry: One of the characteristics present on the faces of drug addicts is an increased facial asymmetry due to drug use. In order to verify this evolution of the face symmetry, a metric is set up based on the work done in \cite{harastani_methamphetamine_2020} and explained in section 2.5.
\end{enumerate}

\subsection{Existing Database}
\begin{figure}[h]
    \centering
    \includegraphics[width = \textwidth]{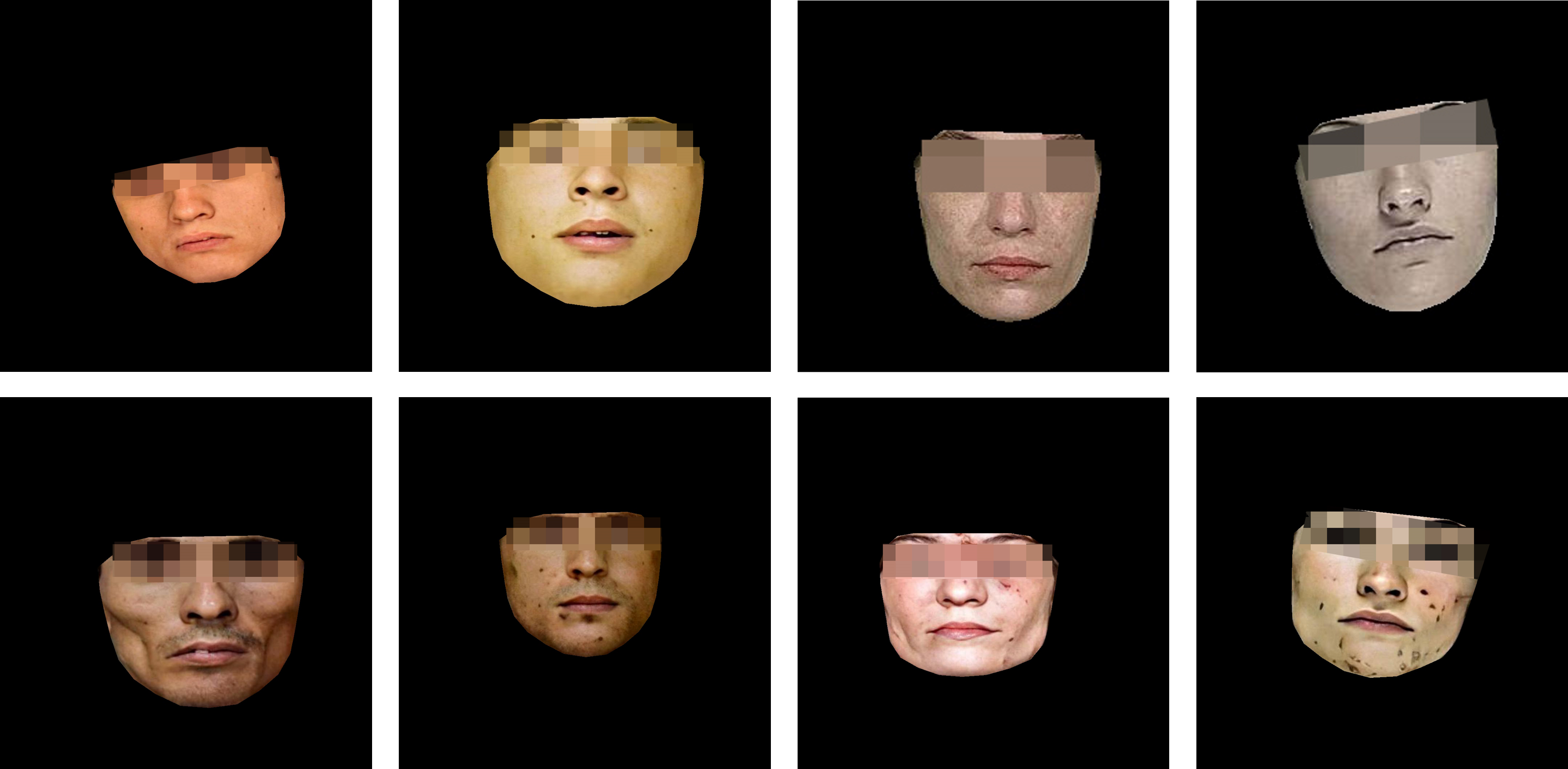}
    \caption{Sample of extracted, blurred and upscaled to $1024 \times 1024$ face images of members of the Illicit Drug Addicts database in early stage of drug abuse in the first row and at a later stage in the second row}
    \label{fig:FOM_samp}
\end{figure}

The database used is Illicit Drug Addicts Dataset. It comes from the Face Of Meth project \cite{faceofmeth}. It contains the faces of 120 people. For each person, there are at least two frontal images. The average resolution is $(572 \times 838)$. This resolution is a limitation because the objective is to generate high resolution images $(1024 \times 1024)$. We therefore proceeded to an upscaling to $(1024 \times 1024)$ using DFDNet \cite{Li_2020_ECCV}, a deep learning model used for the restoration and upscaling of human faces. The dataset has a variety of members belonging to both genders and having an ethnic plurality. A sample of this database is presented in Fig. \ref{fig:FOM_samp}.

\begin{table}[]
\centering
\caption{Characteristic of the real and existing database}
\label{tab:my-table}
\def\arraystretch{1.7}
\begin{tabular}{|c|c|c|c|c|c|}
\hline
\textbf{Size} & \textbf{Type} & \textbf{Average Resolution} & \textbf{Gender} & \textbf{Categories}          & \textbf{Drug Abuse Duration} \\ \hline
242           & Face Images   & 572 x 838                   & Men and Women   & First intake and later stage & from 4 months to 15 years        \\ \hline
\end{tabular}
\end{table}

\subsection{Approach based on genetic algorithms}
\subsubsection{Fundamentals}
\begin{figure}[H]
  \centering
  \subfloat[]{\includegraphics[width=0.47\textwidth]{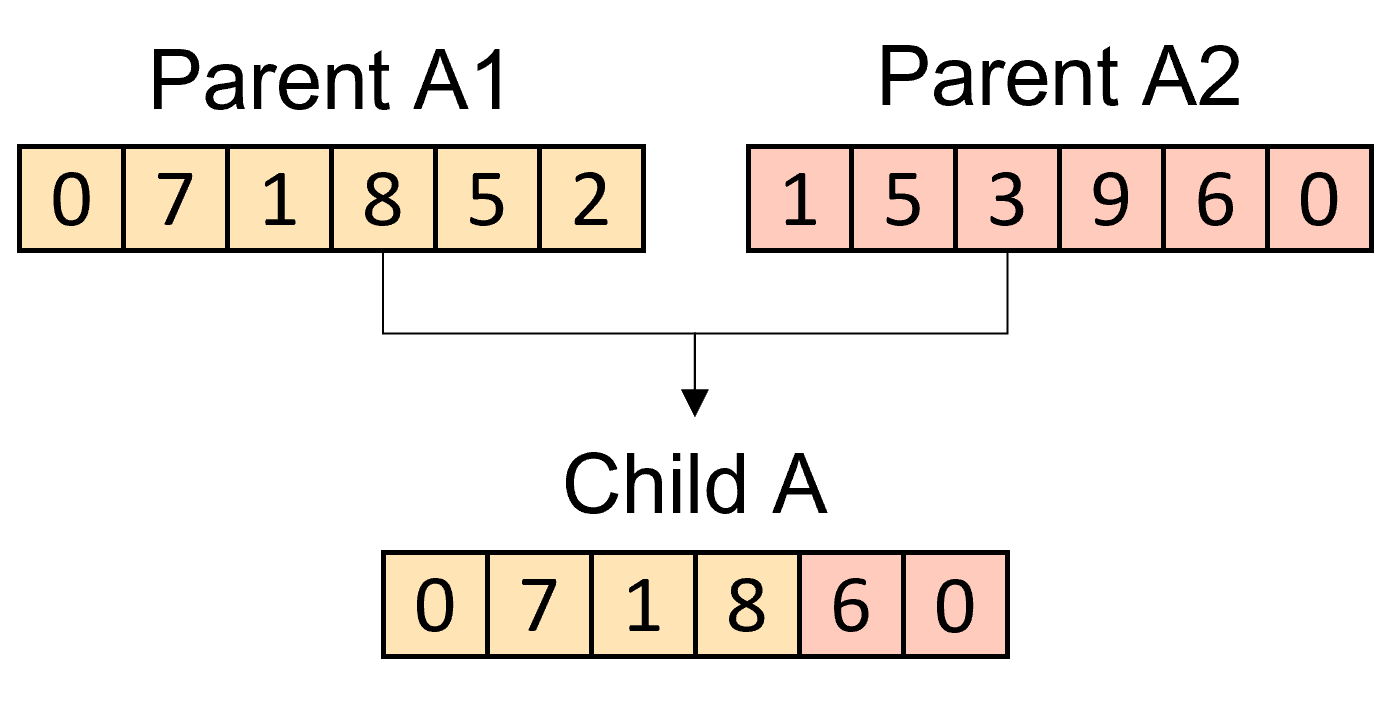}\label{fig:crossover}}
  \hfill
  \subfloat[]{\includegraphics[width=0.47\textwidth]{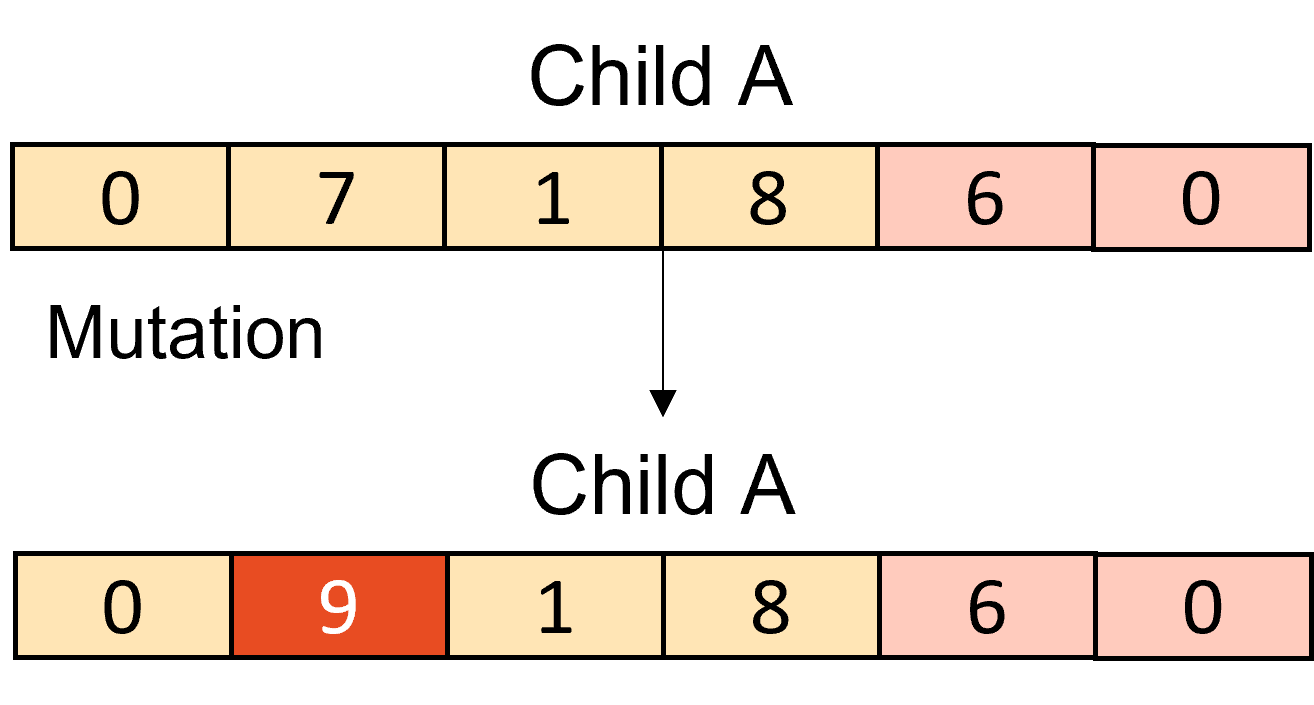}\label{fig:mutation}}
  \caption{Example of genetic algorithms operations on representative data, (a) Crossover operation: The Child A1 takes a first sequence (the 4 firsts genes) belonging to Parent A1 and a second sequence (the 2 last genes) belonging to Parent A2 ; (b) Mutation operation: The Child A1 undergoes a mutation on is second gene which mutes from 7 to 9}
\end{figure}

Genetic algorithms are evolutionary algorithms inspired by process of natural selection \cite{mitchell_introduction_1996}. They are based on 3 points:

\begin{enumerate}
    \item A process of natural selection ensues so that the best adapted chromosomes reproduce more often and contribute more to subsequent generations. It is done at each new generation
    \item During reproduction, the information from the parents, contained in the chromosomes, is combined and mixed to obtain the child chromosomes. This is called cross over or recombination. It is the most common operation. An example of crossover operation is shown in Fig. \ref{fig:crossover} .
    \item The result of the cross can be changed in turn at random. This is called mutation. It changes the properties of an individual at random, appears at any time on anybody. An example of mutation operation is shown in Fig. \ref{fig:mutation}
\end{enumerate}

\subsubsection{Approach}
The method used in this paper is based on the concept of genetic algorithms presented in section 2.2.1. From this concept, a face fusion algorithm \ref{alg:one} has been developed. This algorithm is based on 3 steps wich are repeated at each new generation:

\begin{enumerate}

\item The selection of faces from the faces of drug addicts and the healthy faces generated by GAN.
\item The fusion of faces by crossover or mutation
\item Forgery detection of faces to keep only the faces detected as real 
\end{enumerate}

\begin{algorithm}[H]{
\caption{Database augmentation using face fusion based on genetic algorithm principle}\label{alg:one}
\SetKwInOut{Input}{Input}  
\Input{ $i \gets Facial Drug Abuse Dataset Folder$ \\
    $j \gets GAN Dataset Folder$ \\
    $\alpha \in [0,1]$ \\
    $max\_g \gets Max Number Of Generations$ \\
    $max\_i \gets Max Number Of Images Per Generation$ \\
    $count \gets 0$
}
\textbf{\\}
\For{$g = 0;\ g < max\_g;\ g = g + 1 $}{
        \ForEach{$Image  x \in [i,i+1,i+2,....]$}{%
                \If{$count = max\_i$}{
                    $Break$
                }
            \ForEach{$Image  y \in [j,j+1,j+2,....]$}{%
                \If{$count = max\_i$}{
                    $Break$
                }
            
                \If{ $(FaceDetected(x) \And FaceDetected(y))$ }{%
                    $type \gets WeightedRandom([Crossover,0.95],[Mutation,0.05])$
                    \If{$type = Crossover$}{
                        $im \gets FaceMerge(x*\alpha,y*(1-\alpha))$\newline
                        \If{$ForgeryDetect(im$)}{
                           $count \gets count+1$\newline
                            $SaveImage(im)$
                        }
                    }
                    
                    \If{$type = Mutation$}{
                        $r=Random(0,1,Step=0.1)$\newline
                        $im \gets = FaceMerge(x*r,y*(1-r))$ \newline
                        \If{$ForgeryDetect(im)$}{
                            $count \gets count+1$\newline
                            $SaveImage(im)$
                        }
                    }
                }
            }
        }
    }
}

\KwResult{Augmented Dataset}
\end{algorithm}

\subsection{Evaluation metrics}

In order to ensure the three desired characters for the creation of the database: authenticity, realism, anonymity; three evaluation metrics are used. To ensure the anonymity of the data, a facial recognition algorithm is used. The authenticity of the images is ensured by detecting fakes. And the realistic character is given by the measurement of face asymmetry.

\subsubsection{Facial Recognition}
In biomedical applications, privacy is the main concern when handling sensitive data of individuals. To protect the privacy of the subjects, present in the Illicit Drug Addict dataset, we need a solution to anonymize the faces by not being able to connect the resulting fake faces generated by the genetic algorithm to the first real parents. To overcome this problem a using face$\_$recognition \cite{face-recognition_2020} library in python a facial recognition model was trained on the two original datasets before and after drug abuse, then this model was used in classifying each image in the datasets and keeping all the images classified as unknown, to eliminate any possibility of relating a fake image to its real parent and anonymizing all the faces present in the dataset.

\begin{figure}[H]
    \centering
    \includegraphics[width=\textwidth]{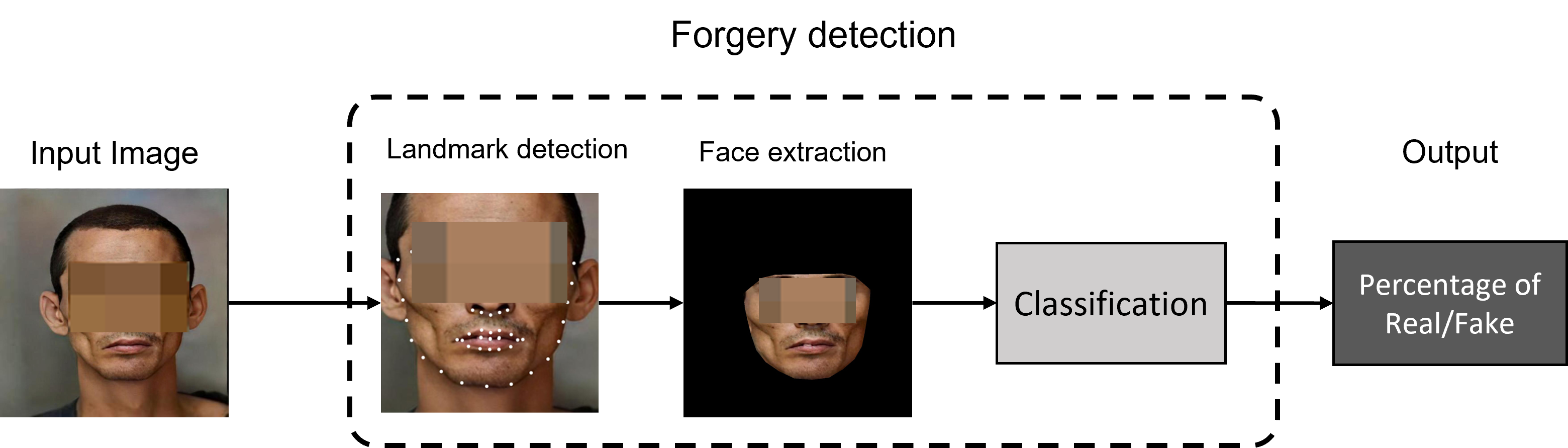}
    \caption{Forgery detection workflow diagram. A landmark detection on the input image is processed to extract the face, then a classification is done to give as output a percentage of Real/Fake of the image}
    \label{fig:forgery}
\end{figure}

\subsubsection{Forgery Detection}
In genetic algorithm, the selection step requires to have a metric to decide whether to keep or reject a sample, the process is shown in Fig. \ref{fig:forgery}. In our case, we are working with images which provides a challenge in quantifying these images into a metric usable in selecting the best samples. To overcome this problem, in each iteration of selection we used a forgery detection model to keep all the fake images that look very realistic with no signs of fusion and bad alignment. The model we used XceptionNet \cite{chollet_xception:_2016} was trained on FaceForensics++ dataset \cite{rossler_faceforensics++:_2019} containing 1000 videos of facial forgeries and it can be used in image and video classification into either fake or real (real/fake confidence percentage).

\subsubsection{Facial asymmetry}
\begin{figure}[H]
    \centering
    \includegraphics[width = 0.42\textwidth]{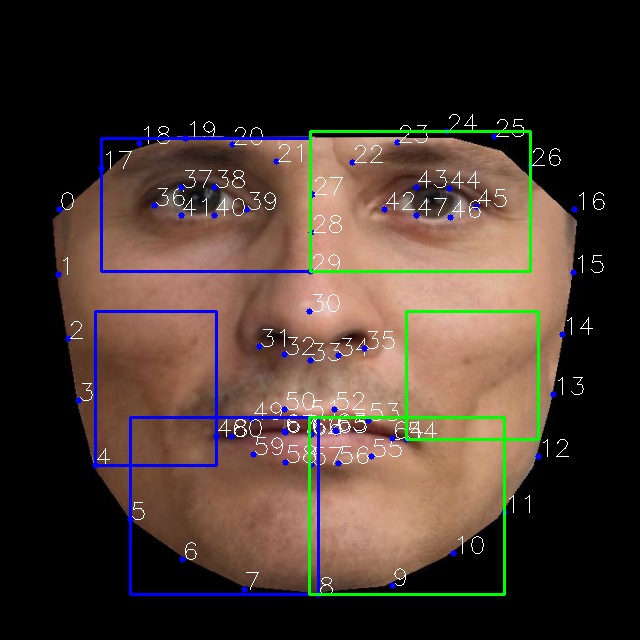}
    \caption{Face landmarks coordinates and regions of interests}
    \label{fig:image}
\end{figure}

Drug abuse was proven to cause facial changes including aging and facial asymmetry directly affecting results of facial recognition. To prove that our generated datasets conserve and present correct facial traits of drug abuse, we did a facial asymmetry analysis \cite{harastani_methamphetamine_2020} on all the images. The method used is presented in Table \ref{tab:coord_landmark} and in Fig. \ref{fig:image}. The following steps were used to reach a metric used in the assessment of facial asymmetry:

\begin{enumerate}
    \item Face Detection and face extraction,
    \item Landmarks detection using DLIB 68 landmarks detector to apply facial alignment and resizing all images to equal resolution,
    \item Extracting regions of interests using DLIB facial landmarks detection:
    \begin{itemize}
        \item Left and right eye areas
        \item Left and right cheek areas
        \item Left and right mouth areas
    \end{itemize}
    \item Structural similarity index (SSIM) was used to compare each two corresponding areas resulting in a similarity score between 0 and 1, where 1 means both images are identical
    \item Finally, we averaged the three scores into one score ranging between 0 and 1 where 1 means both areas are identical
\end{enumerate}

\begin{table}[H]
\centering
    \caption{Landmark coordinates used for cropping regions of interest}
\def\arraystretch{1.4}
    \begin{tabular}{|c|c|c|}
        \hline
        \textbf{ROI} & \textbf{Point 1} & \textbf{Point 2} \\
        \hline
        LEFT EYE & $(X_{17}, Y_{19})$ & $(X_{29}, Y_{29})$ \\
        \hline
        RIGHT EYE & $(X_{29}, Y_{24})$ & $(X_{26}, Y_{29})$ \\
        \hline
        LEFT CHEEK & $(X_{4}, Y_{30})$ & $(X_{48}, Y_{4})$ \\
        \hline
        RIGHT CHEEK & $(X_{54}, Y_{30})$ & $(X_{12}, Y_{54})$ \\
        \hline
        LEFT MOUTH & $(X_{5}, Y_{51})$ & $(X_{8}, Y_{8})$ \\
        \hline
        RIGHT MOUTH & $(X_{51}, Y_{51})$ & $(X_{11}, Y_{8})$ \\
        \hline
    \end{tabular}
    \label{tab:coord_landmark}

\end{table}

\paragraph{Structural Similarity}
Structural similarity \cite{wang_multiscale_2003} is mainly used to assess the quality degradation of pictures after compression compared to the original one. SSIM can be also used for measuring the similarity between two images . the function shown in equation (\ref{SSMIequ}) takes two images x and y on the input having the same sizes, and on the output it returns a similarity score  between x and y in the range [0,1] where 1 means that x=y (identical images).

\begin{equation}
    SSIM(x,y) = \frac{(2 \mu_x \mu_y + C_1)(2 \sigma_{xy} + C_2)}{(\mu^2_x + \mu^2_y + C_1)(\sigma^2_x + \sigma^2_y + C_2)}
    \label{SSMIequ}
\end{equation}

Where:

\begin{enumerate}
    \item $\mu_x$ and $\mu_y$ are the average of pixel values in images $x$ and $y$
    \item $\sigma^2_x$ and $\sigma^2_y$ are the variance in images $x$ and $y$
    \item $\sigma_{xy}$ is the covariance in images $x$ and $y$
\end{enumerate}

\section{Results and Discussion}
The objective of this paper is to provide a method to augment a reduced complex database. A reduced database contains few data, in our case about 200 images. This is not enough to use it as input for GANs. And the complex character of the data comes from the presence of many features on the face of the drug addicts which must be reproduced. Using the method described above, we managed to generate a database containing 5 generations of images for each value of $\alpha$ from 0 to 1 (the coefficient $\alpha$ represents the percentage of drugged image used in the fusion) process for pre-drug and post-drug faces. Each generation contains 300 face images labeled as true during the forgery detection for both before and after images. Characteristics of this generated database are shown in Table \ref{tab:charac_result}. 

\begin{table}[H]
\centering
\caption{Characteristics of the augmented database}
\label{tab:charac_result}
\resizebox{1\columnwidth}{!}{%
\begin{tabular}{|c|c|c|c|c|c|c|}
\hline
\textbf{Size} &
  \textbf{\begin{tabular}[c]{@{}c@{}}Numeber of\\ Genrations\end{tabular}} &
  \textbf{Type} &
  \textbf{Resolution} &
  \textbf{Gender} &
  \textbf{\begin{tabular}[c]{@{}c@{}}Drug Abuse \\ Duration\end{tabular}}  \\ \hline
\begin{tabular}[c]{@{}c@{}}3000 images for each alpha\\ value ranging from 0.0 to 1.0,\\ reaching a total of 33000 images\end{tabular} &
  5 &
  Face Images &
  1024x1024 &
  Men and Women &
  from 4 months to 15 years  \\ \hline
\end{tabular}%
}
\end{table}


\begin{figure}[H]
    \centering
    \includegraphics[width = \textwidth]{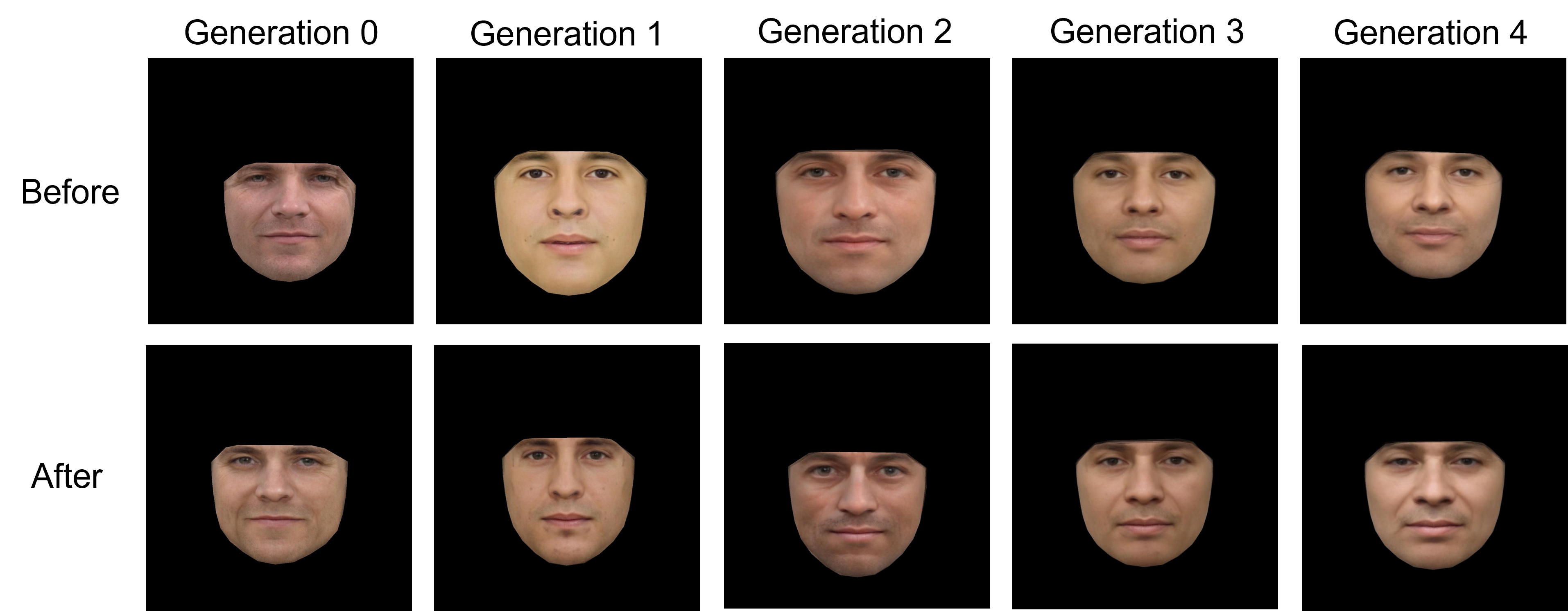}
    \caption{Samples of preserved generated faces for an $\alpha = 0,5$ and for five different generations}
    \label{fig:sample_gen}
\end{figure}

These images are therefore the images kept during the whole process. A sample of this result can be seen in Fig. \ref{fig:sample_gen} for an $\alpha = 0.5$. On this figure features can be visually observed that we associate with drug addicts, namely the dark circles that are accentuated between the before and after images. Also, the weight loss on some patients is visible. Finally, the presence of after-effects on the skin is increased in this before/after comparison. Finally, the presence of after-effects on the skin is increased in this before/after comparison. However, it can be observed that over the generations, a smoothing of the skin becomes more pronounced. This smoothing is expected because of the fusion between the two input images. We can therefore say that the smoothing increases with the number of generations, the images thus generated will be more difficult to accept as authentic images. 

\begin{figure}[H]
    \centering
    \includegraphics[width = \textwidth]{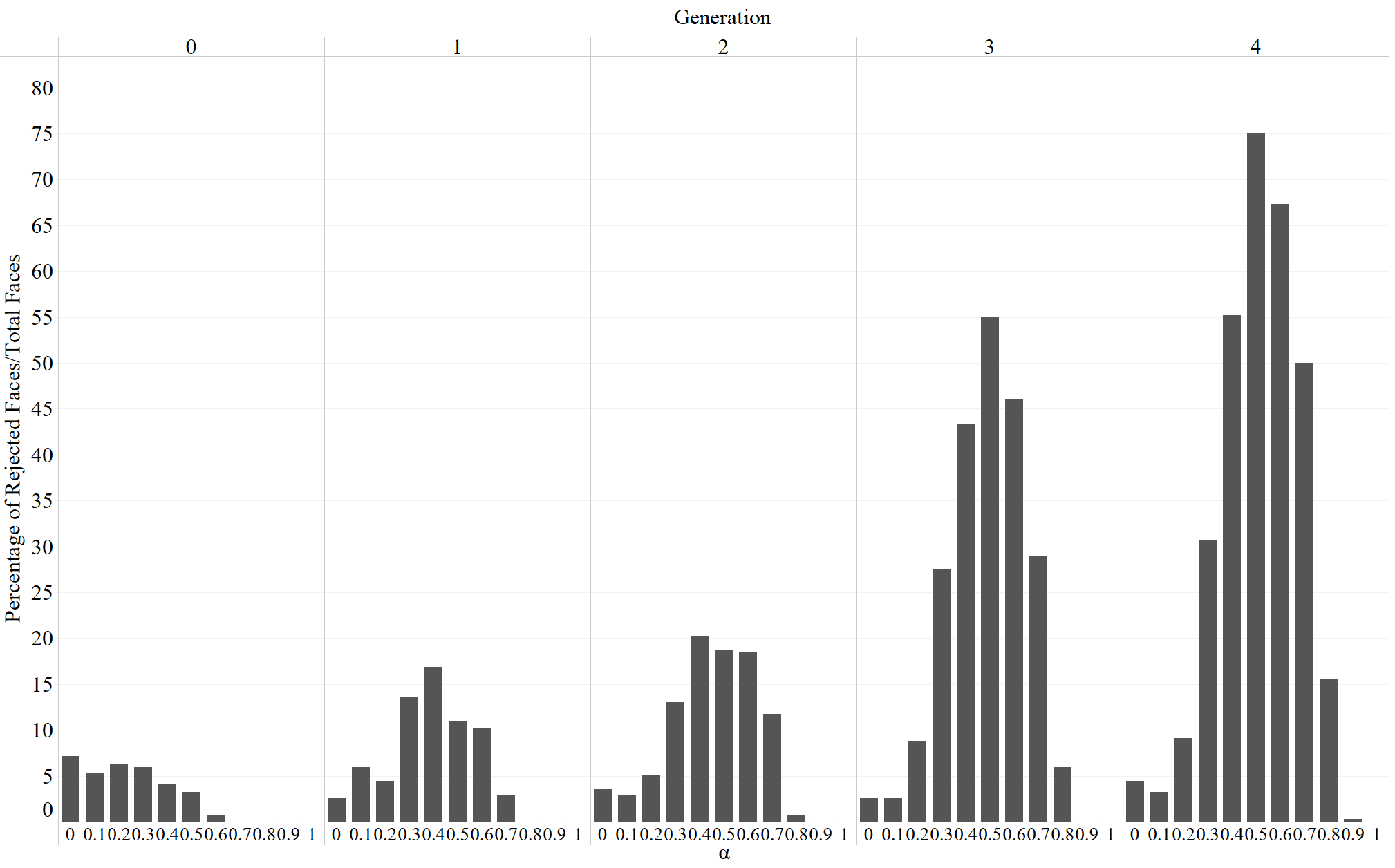}
    \caption{Evolution of the number of images rejected, labelled as fake, over the number of generation and for $\alpha$ from 0 to 1. Number of images rejected is increasing with the number of generation. At each generation, a symmetry is observed around $\alpha = 0.5$}
    \label{fig:rejected}
\end{figure}

It then comes to the image rejection rate. This rate is expected to increase with the number of generations but also that within a same generation this rate is more important for values of $\alpha$ close to the average, here $\alpha = 0.5$. Fig. \ref{fig:rejected} shows the evolution of this rate as a function of the generations and as a function of the value of $\alpha$. First, we can observe that as the number of generations increases, the number of rejected images also increases. The global approach is then in accordance with the expectations. By observing, now, on each generation, a symmetry is noticed on the rate of rejection of the images compared to the average. This behavior is similar to that of a Gaussian distribution. When $\alpha$ is close to the extreme values (here 0 and 1), the rejection rate decreases, on the contrary, when we approach the average value, this rate increases. It increases the more the generation concerned is large until reach $75\%$ for an $\alpha = 0.5$ at the last generation. It should be noted that even if this value is high, the apparent images in each generation are it is images kept is therefore authentic.

Let now consider the anonymity of the generated data. Increasing of the anonymity with $\alpha$ is expected and this regardless of the generation concerned. By observing Fig. \ref{fig:facerec_rate}, we can confirm the hypothesis announced previously. This result is not influenced by the images of the faces before or after. The larger the observed generation is, the more this rate decreases for each value of $\alpha$ and ends up stabilizing at $100\%$ of identification for values of $\alpha$ close to 1. This is due to the fact that our face recognition is based on a comparison between the characteristics of the generated image and the characteristics of the images belonging to the initial database.

\begin{figure}[H]
  \centering
  \subfloat[]{\includegraphics[width= \textwidth]{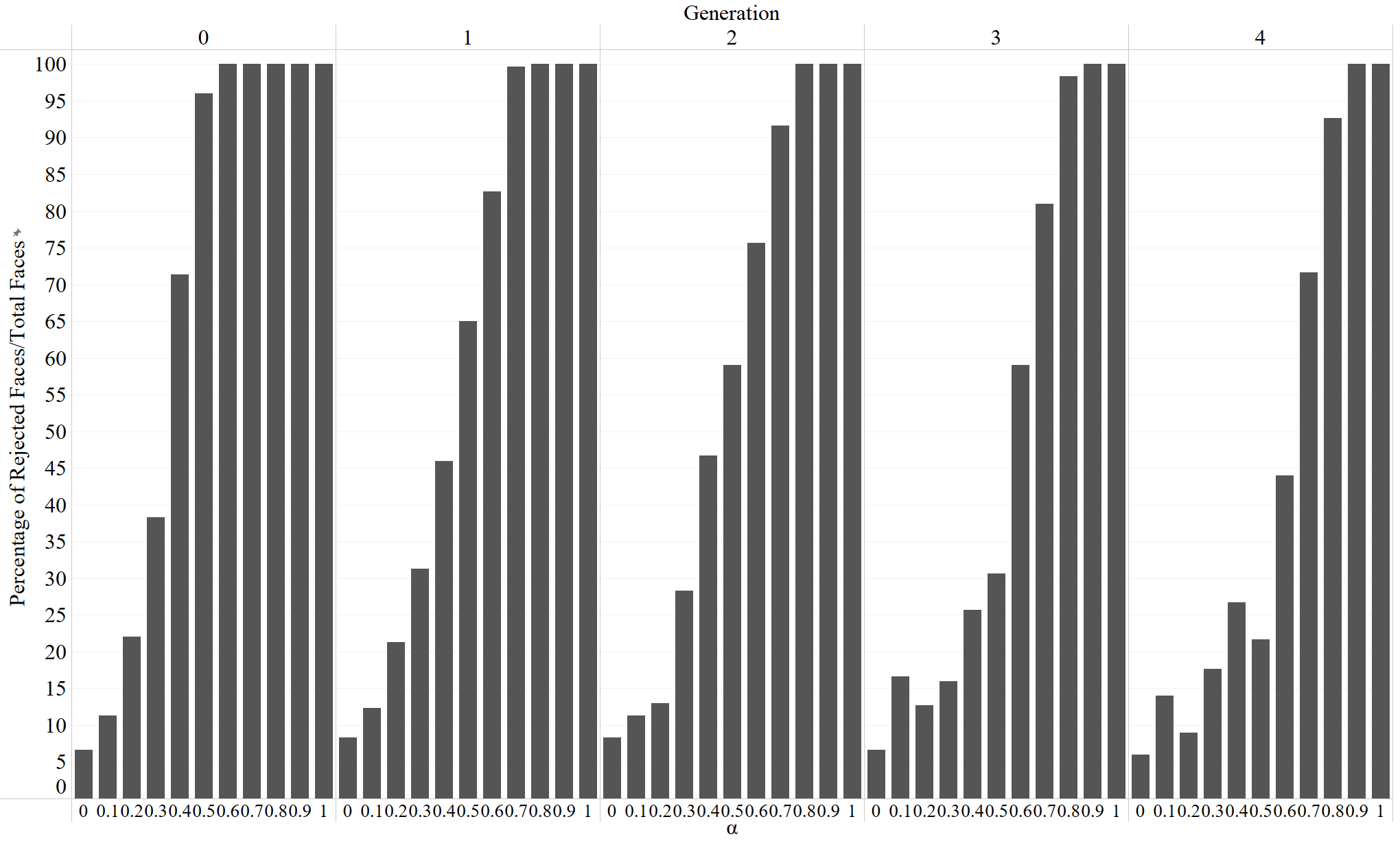}\label{fig:facerec_b}}
  \label{fig:facerec_rate}
\end{figure}

\begin{figure}[H]
  \centering
  \subfloat[]{\includegraphics[width= \textwidth]{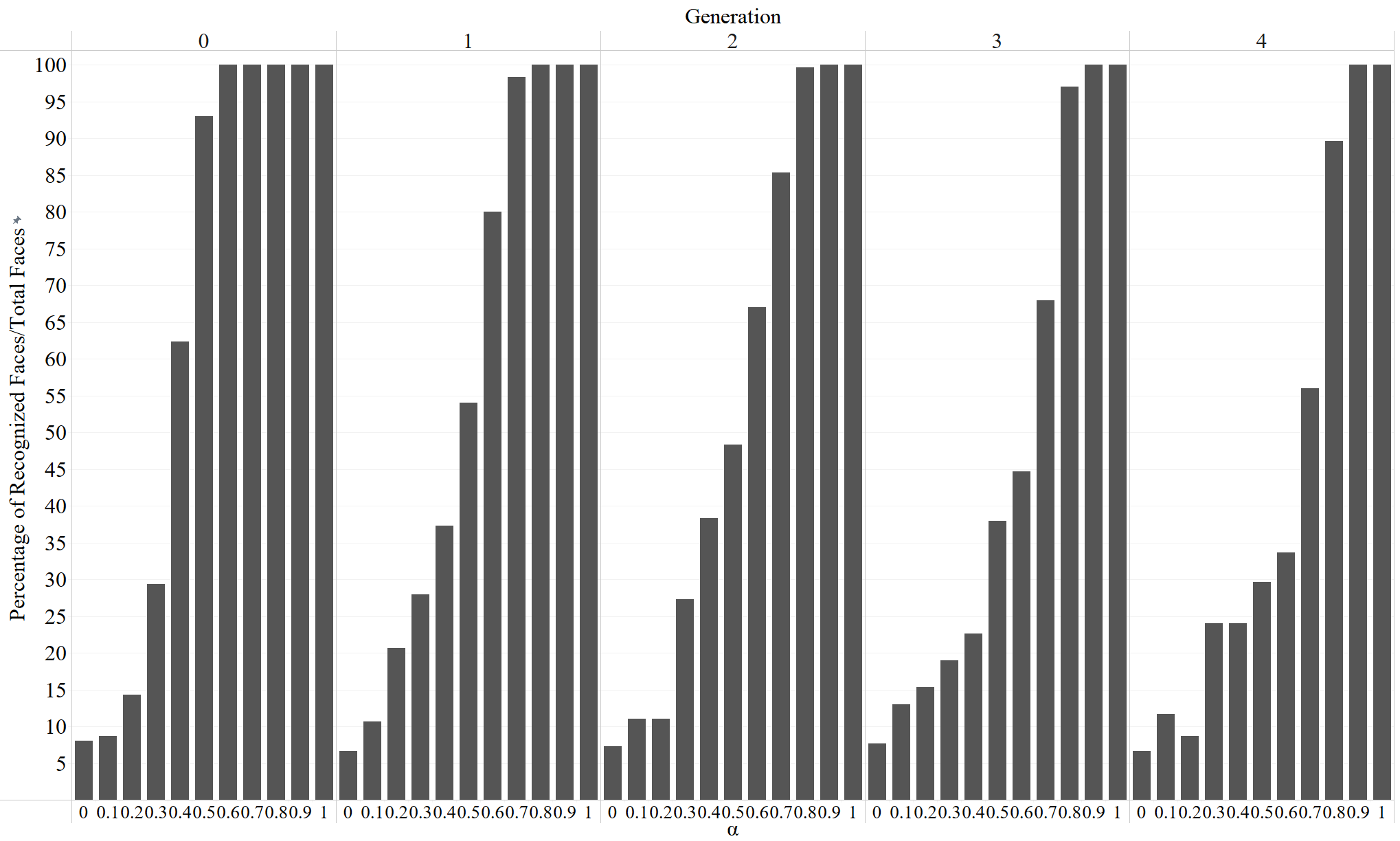}\label{fig:facerec_a}}
  \caption{Evolution of the number of images recognized over the number of generation and for $\alpha$ from 0 to 1 for, (a) images before and, (b) images after}
  \label{fig:facerec_rate}
\end{figure}

\begin{figure}[H]
  \centering
  \subfloat[]{\includegraphics[width=0.48\textwidth]{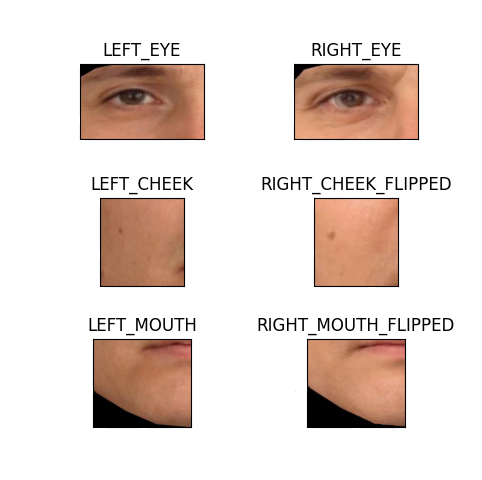}\label{fig:asymmetry_b}}
  \hfill
  \subfloat[]{\includegraphics[width=0.48\textwidth]{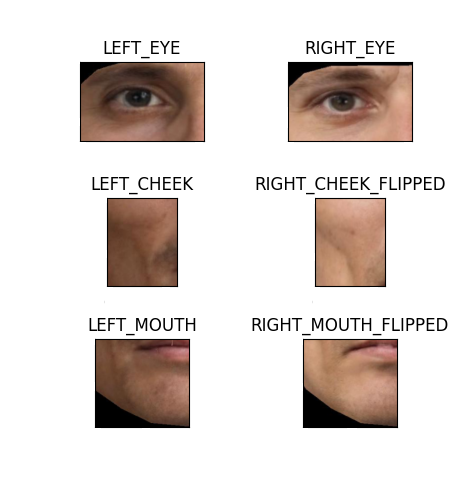}\label{fig:asymmetry_a}}
  \caption{Example of the evolution of face symmetry on left and right parts of face (eyes, cheeks, mouth) for images before and after with $\alpha = 0,5$}
  \label{fig:asymmetry}
\end{figure}

We said earlier that one of the representative factors in drug addicts was the evolution of facial asymmetry in an increased manner. In order to verify that the generated data respected this observation, a measurement of the symmetry of the generated face of a person before and after was made for each image contained in the generated database. These symmetry measurements allowed to expose the evolution of the asymmetry in the generated drug addicts. Fig. \ref{fig:asymmetry} shows an example from our data. Table \ref{tab:asymetry} shows the values obtained from the face symmetry measurements. A value close to $100\%$ means that the analyzed part of the face is symmetrical. This measurement is performed on the eyes, the cheeks and the mouth. We can see that this symmetry decreases on the face after drug use. This confirms the hypothesis formulated above, the asymmetry of the face is accentuated by the drug consumption.

\begin{table}[H]
    \centering
        \caption{Results of the symmetry measurement for asymmetry analysis and evolution of asymmetry between before and after. The symmetry between before and after is decreasing (values before are closer to $100\%$), this mean that the asymmetry is increasing after drug use}
        \def\arraystretch{1.5}
    \begin{tabular}{|c|c|c|}
        \hline
        \textbf{REGION} & \textbf{Before} & \textbf{After} \\
        \hline
        EYES & 66.5\% & 46.4\% \\
        \hline
        CHEEKS & 83.3\% & 69.9\% \\
        \hline
        MOUTH & 77.9\% & 69.2\% \\
        \hline
    \end{tabular}

    \label{tab:asymetry}
\end{table}

\section{Conclusion}
The focus of this paper is to present a method to overcome the problem of lacking large enough datasets to be used in deep learning or other applications. One example is facial drug abuse datasets that are not available publicly in large numbers due to privacy reasons. Our objective is to provide a method to augment datasets with fake but highly realistic images that can replace real images in deep learning applications. To tackle this problem, we propose using genetic algorithms to generate fake images while keeping the best samples after each generation. We started with around 200 images of facial drug abuse split into before and after drug abuse, this dataset presents a low number of images that can not be used in any application, therefor we generated 73 healthy face images StyleGAN2  to be used in the reproduction with facial drug abuse images. Many preprocessing steps were applied to these two datasets to ensure a good result on the output: facial alignment, face extraction, image resizing. In the genetic algorithm, fusion was applied to images from both GAN and Real facial drug abuse on each generation, and in the selection process a forgery detection model was used, classifying images as either real or fake, fake images were discarded and only images classified as real proceeded to the next generation. To guarantee the anonymity of the real parents and to avoid any privacy concern we trained a facial recognition model on our real facial drug abuse images and ran it on our generated dataset of fake facial abuse images.An analysis was done to show the preservation of facial drug abuse traits through generation, one example is the increase in facial asymmetry, results showed that our genetic algorithm preserved this trait through generations.

\end{document}